\documentclass[runningheads]{llncs}
\usepackage[T1]{fontenc}
\usepackage{graphicx}
\usepackage{booktabs}

\usepackage{amsmath}
\usepackage{calc}
\usepackage{amssymb}
\usepackage{amstext}

\usepackage[hidelinks]{hyperref}
\usepackage{color}

\begin{document}

\title{Distilling Local Texture Features for Colorectal Tissue Classification in Low Data Regimes}

\titlerunning{Colorectal Tissue Classification}
%
\author{Dmitry Demidov$^*$ 
\and
Roba Al Majzoub$^*$ 
\and
Amandeep Kumar
\and
Fahad Khan 
}
%
%
\institute{
Mohamed Bin Zayed University of Artificial Intelligence, Abu Dhabi, UAE \email{\{dmitry.demidov, roba.majzoub, amandeep.kumar, fahad.khan\}@mbzuai.ac.ae}}

\maketitle              

\begin{abstract}
    Multi-class colorectal tissue classification is a challenging problem that is typically addressed in a setting, where it is assumed that ample amounts of training data is available. However, manual annotation of fine-grained colorectal tissue samples of multiple classes, especially the rare ones like stromal tumor and anal cancer is laborious and expensive. To address this, we propose a knowledge distillation-based approach, named KD-CTCNet, that effectively captures local texture information from few tissue samples, through a distillation loss, to improve the standard CNN features. The resulting enriched feature representation achieves improved classification performance specifically in low data regimes. Extensive experiments on two public datasets of colorectal tissues reveal the merits of the proposed contributions, with a consistent gain achieved over different approaches across low data settings. 
    The code and models are publicly available on \href{https://github.com/demidovd98/med_lowdata}{\textit{GitHub}}.

\end{abstract}

\def\thefootnote{*}\footnotetext{These authors contributed equally to this work}\def\thefootnote{\arabic{footnote}}

\section{Introduction}
Colorectal cancer (CRC) remains a prominent global health concern, claiming over 10 million lives in 2020 \cite{who} and ranking as the second leading cause of death. In response, significant research efforts have been dedicated to facilitating early diagnosis and treatment through histopathology tissue data classification. The automatic classification of such images in general presents a complex and vital task, enabling precise quantitative analysis of tumor tissue characteristics and contributing to enhanced medical decision-making.

Colorectal tissue classification is typically solved in a setting that assumes the availability of sufficient amounts of labelled training samples \cite{kather2016multi,kumar2023crccn,tsai2020deep}. Therefore, most existing works on colorectal tissue classification follow this setting based on datasets with hundreds to thousands of annotated training images.  However, this is not the case for rare instances such as gastrointestinal stromal tumors, anal cancers, and other uncommon CRC syndromes, where data availability is limited. Further, manual annotation of such fine-grained colorectal tissue data of multiple classes is laborious and expensive. To reduce the dependency on the training data, 
few works have explored other settings such as self-supervised~\cite{saillard2021self}, semi-supervised \cite{bakht2021colorectal}, and few-shot learning \cite{shakeri2022fhist}. However, both self-supervised and semi-supervised colorectal tissue classification scenarios assume the availability of unlabeled colorectal tissue training data, whereas the few-shot counterpart setting relies on a relatively larger amount of labelled training samples belonging to base classes for general feature learning. In this work, we explore the problem of colorectal tissue classification 
in a more challenging setting, termed \textit{low data regime}. In this setting, the number of labelled training samples for all the classes is significantly reduced, i.e., 1\% to 10\% \textit{per} category.

The aforementioned challenging setting of low data regimes has been recently studied \cite{shu2022improving} in the context of natural image classification. 
The authors proposed an approach that improves the feature representation of standard vanilla convolutional neural networks (CNNs) by having a second branch that uses off-the-shelf class activation mapping (CAM) technique \cite{Grad_CAM} to obtain heat maps of gradient-based feature class activations. The feature-level outputs of both branches are trained to match each other, which aids to refine the standard features of the pre-trained CNN in the low-data regimes. While effective on natural images having distinct structures of \textit{semantic objects}, our empirical study shows that such an approach is sub-optimal for colorectal tissue classification likely due to CAM being ineffective in capturing \textit{texture} feature representations in colorectal tissues. Therefore, we look into an alternative approach to address the problem of colorectal tissue classification in low data regimes.

When designing such an approach, effectively encoding local texture information is crucial to obtain improved classification performance \cite{KHAN201516}. One approach is to utilize a pre-trained model and fine-tune it to encode data-specific local textures. This has been previously endorsed in~\cite{chen2019general,Sun_Cholakkal_Khan_Khan_Shao_2020,tsai2020deep}. Model fine-tuning in standard settings with ample training samples allows it to encode local texture information. However, this strategy may struggle with extremely scarce training data. A straightforward alternative is random cropping, where smaller regions of the input image are resized for data augmentation to capture local texture patterns~\cite{caron2021emerging}. However, such data augmentations prove to be sub-optimal due to the large pixel-level input variations and the artificial generation of a large portion of the pixels due to data extrapolation. To address this issue,  we adopt knowledge distillation, transferring knowledge between models without performance sacrifice. This approach has been previously used for both natural and medical image classification~\cite{caron2021emerging,javed2023knowledge}. In this paper, we utilize all three approaches combined to mitigate the effect of data scarcity on model training and allow the model to learn better representative features even in low data regimes.

\noindent\textbf{Contributions:} We propose a knowledge distillation-based approach, named KD-CTCNet, to address the problem of colorectal tissue classification in low data regimes. Our proposed KD-CTCNet strives to improve the feature representation of the standard fine-tuned CNN by explicitly capturing the inherent local texture information from very few colorectal tissue samples. Our KD-CTCNet consists of a standard branch with a conventional CNN stream and a local image branch that performs local image sampling to encode local texture information. We further utilize a knowledge distillation technique, where the standard global branch serves as a teacher and the local image branch acts as a student. The corresponding output logits of the two branches are compared through a distillation loss to obtain enriched feature representations that are specifically effective for low data regimes. Extensive experiments conducted on two histopathological datasets of CRC including eight and nine different types of tissues respectively, reveal the merits of our proposed contributions. Our KD-CTCNet consistently outperforms standard pre-trained and fine-tuned CNN approaches on a variety of low data settings. 

\begin{figure}[!t]
    \caption{Samples from Kather-2016 (top) and Kather-2019 (bottom) datasets.}
    \includegraphics[width=.125\textwidth]{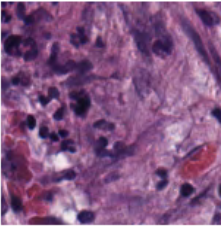}\hfill
    \includegraphics[width=.125\textwidth]{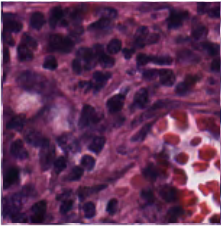}\hfill
    \includegraphics[width=.125\textwidth]{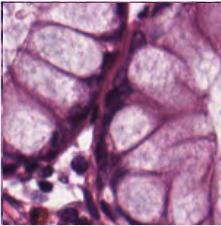}\hfill
    \includegraphics[width=.125\textwidth]{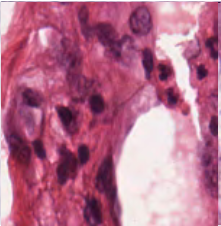}\hfill\hfill
    \includegraphics[width=.125\textwidth]{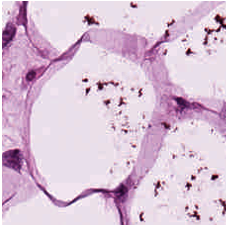}\hfill
    \includegraphics[width=.125\textwidth]{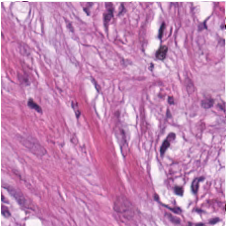}\hfill
    \includegraphics[width=.125\textwidth]{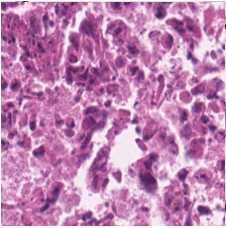}\hfill
    \includegraphics[width=.125\textwidth]{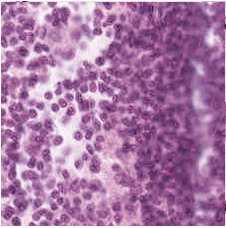}
    \\[\smallskipamount]
    \includegraphics[width=.111\textwidth]{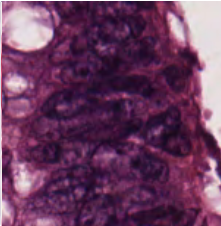}\hfill
    \includegraphics[width=.111\textwidth]{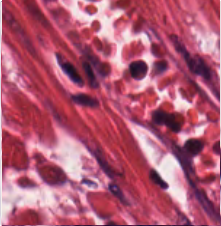}\hfill
    \includegraphics[width=.111\textwidth]{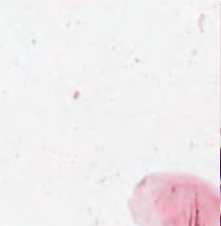}\hfill
    \includegraphics[width=.111\textwidth]{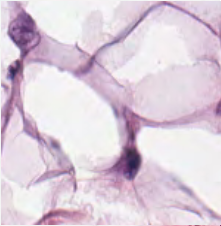}\hfill\hfill
    \includegraphics[width=.111\textwidth]{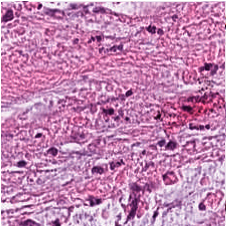}\hfill
    \includegraphics[width=.111\textwidth]{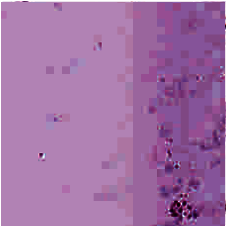}\hfill
    \includegraphics[width=.111\textwidth]{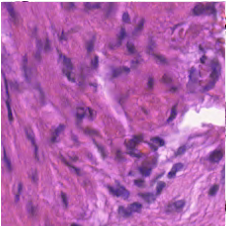}\hfill
    \includegraphics[width=.111\textwidth]{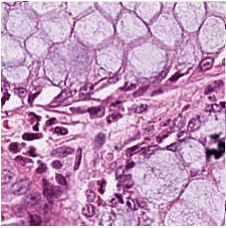}\hfill
    \includegraphics[width=.111\textwidth]{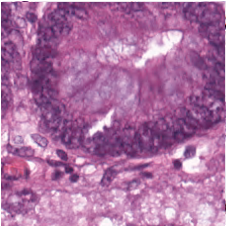}
    
    \label{fig:KatherDatasets}
\end{figure}

\begin{table}[!t]\centering
\caption{Number of images per class for train sets of the Kather-2016 dataset in our low data sampling strategy.}
\begin{tabular}[\textwidth]{lccccccccccc}
    \toprule
    \bfseries Percentage & \textbf{1$\%$} & \textbf{3$\%$} & \textbf{5$\%$} & \textbf{10$\%$} & \textbf{20$\%$} & \textbf{30$\%$} & \textbf{40$\%$} & \textbf{50$\%$} & \textbf{75$\%$} & \textbf{100$\%$} \\
    
    \midrule
    Samples/class &  3 & 9 & 15 & 30 & 62 & 93 & 124 & 156 & 234 & 312\\
    
    \bottomrule
\end{tabular}
\label{tab:sampling}
\end{table}

\noindent\textbf{Datasets:} Our main experiments are conducted on \textbf{Kather-2016}, a dataset of CRC images introduced by \cite{kather2016multi}. Image patches of size $150\times150$ are extracted from 10 anonymized whole slide images (WSI) obtained from Mannheim University Medical Center, Heidelberg University, Mannheim, Germany. The dataset includes 5,000 images equally distributed among eight classes, with 625 images per class. To further validate our approach we conduct additional experiments on \textbf{Kather-2019}, the NCT-CRC-HE-100K dataset~\cite{kather2019predicting}. This publicly available dataset contains nearly 100,000 H\&E stained patches of size $224\times224$. These patches are extracted from Whole Slide Images (WSI) of 86 patients sourced from NCT Biobank and the UMM pathology archive. The dataset comprises 9 classes, and sample images are shown in Fig~\ref{fig:KatherDatasets}.\\
Although these datasets were previously utilized for histopathology classification~\cite{dabass2022convolution,anju2022tissue}, most of the performed research utilizes either all the data or a very large percentage of it, which does not reflect real-life scenarios for many diseases.\\
\noindent\textbf{Sampling Strategy.} 
To address low data regimes, we employ a sampling approach on the dataset, mimicking limited data availability for rare cases and diseases. Each class is divided into 50\% training and 50\% testing sets. Subsequently, we randomly sample various percentages of the training data (ranging from 1\% to 100\% of the train set with an equal number of images sampled per class) for model training. This sampling of training images is done three times randomly and the resulting model is tested on the same test set originally split from the overall dataset. Table~\ref{tab:sampling} presents the number of images for each percentage. To further validate our model's performance, we first randomly sample 625 images from each class of the Kather-2019 dataset (to match the number of images in the Kather-2016 dataset) and then repeat the sampling scheme.

\section{Method}

\subsection{Baseline Framework}

\noindent\textbf{Transfer learning.}
A ResNet model~\cite{he2015deep}, pre-trained on ImageNet \cite{ImageNet} is fine-tuned on the downstream colorectal tissue classification task. Typically, fine-tuning allows the transfer of generic features learned on large datasets across similar tasks for better model initialization in cases of small training datasets. Here, we are given 
source domain data $D_S$ and target-domain data $D_T$, each consisting of a feature space $\chi$ and probability distribution $P(X)$, where $X = \{x_1, ..., x_n\} \in \chi$. 
Therefore, the task to solve is a prediction of the corresponding label, $f(x')$, of a new instance $x'$ using label space $Y$ and an objective function $f(\cdot)$, which should be learned from the labelled training data of pairs $\{x_i, y_i\}$, where $x_i \in X$ and $y_i \in Y$ represent samples and their assigned labels.

\noindent\textbf{Limitations.}
The current scheme is sub-optimal due to two reasons. First, the scarcity of training samples in low data regimes leads to model overfitting. Second, previous works in texture recognition literature \cite{Lazebnik05,SIMON20201680} have shown that local feature representations are more effective in encoding distinct texture patterns found in colorectal histopathology tissue samples. We empirically show that fine-tuning an ImageNet pre-trained model on colorectal tissue data in a low data regime is sub-optimal in capturing local texture patterns due to limited training data. Consequently, standard transfer learning is relatively ineffective in this context. To address these challenges, we propose an approach that prioritizes capturing local texture patterns to improve feature representations in low-data settings.

\subsection{Local-Global feature Enrichment via Knowledge Distillation}

\begin{figure}[!t]
    \includegraphics[width=\textwidth]{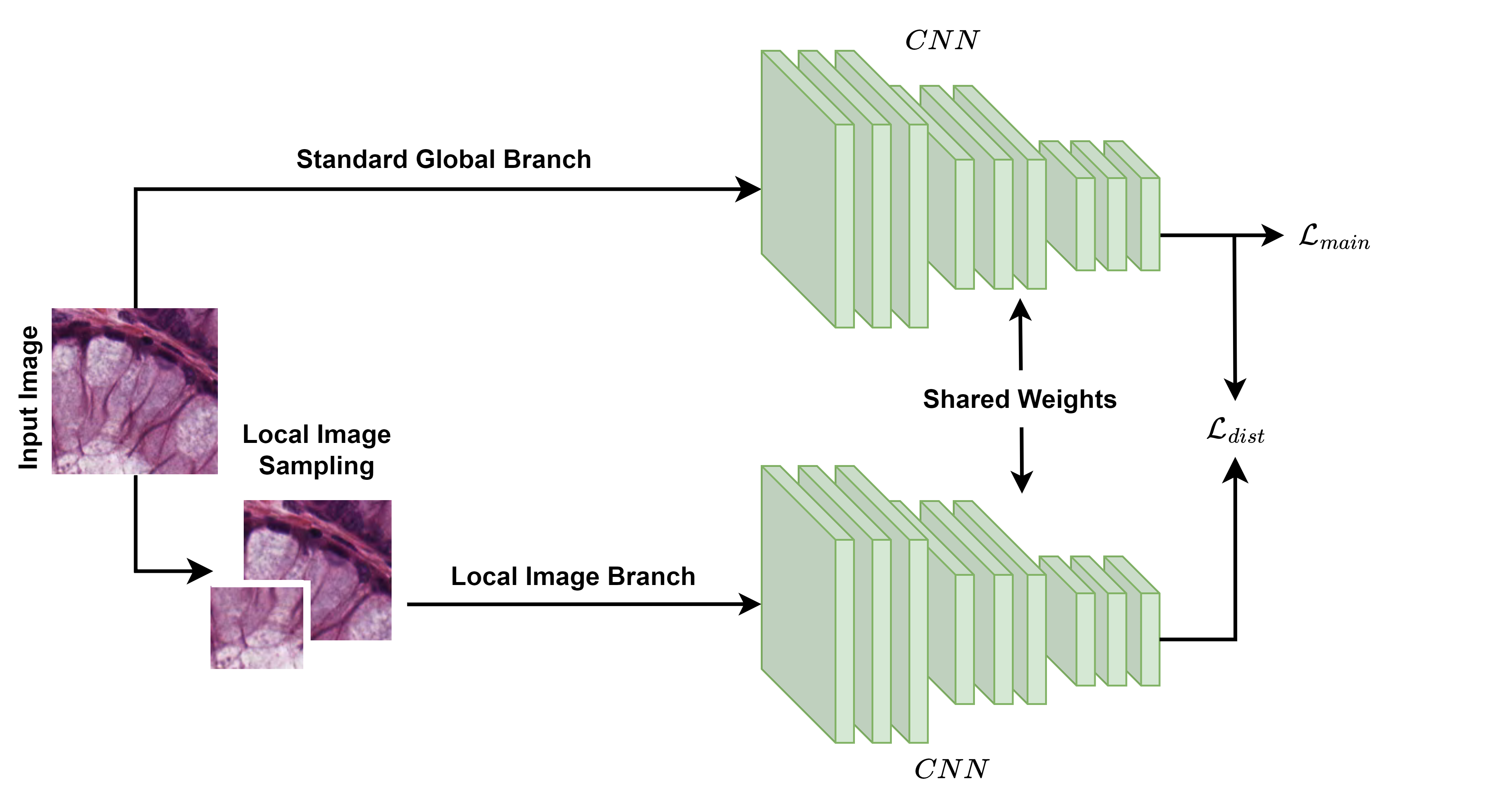}
    \caption{Overview of our proposed knowledge distillation-based framework, named KD-CTCNet, for CRC tissue classification in low data regimes.
    The framework strives to obtain enriched feature representations by explicitly capturing the inherent local texture patterns within the CRC data. 
    The proposed KD-CTCNet comprises a conventional CNN stream (top branch) encoding features from the full image content with a standard cross-entropy loss ($\cal L$$_{main}$) and a parallel branch that is specifically designed to capture local texture feature representations by performing local image sampling. Both branches share the weights and their corresponding output logits are compared using a self-distillation loss ($\cal L$$_{dist}$). The resulting enriched feature representations are beneficial to obtain improved classification performance, especially in low data regimes.
    } 
    \label{fig:architecture}
\end{figure}

\noindent\textbf{Overall Architecture.}
Fig. \ref{fig:architecture} shows an overview of our proposed architecture of a knowledge distillation-based framework, named KD-CTCNet, for CRC tissue classification in low data regimes. The proposed KD-CTCNet comprises two branches: a standard global branch and a local image branch. The standard global branch, a pre-trained ResNet model, takes the entire image as input and utilizes a standard cross-entropy loss. On the other hand, the local image branch first performs local image sampling by randomly cropping different-sized regions from the input image. The two branches in our framework use the shared weights and their corresponding output logits are compared using a self-distillation loss. 

\noindent\textbf{Standard Global Branch.}
This branch represents the standard CNN stream, which follows the conventional training procedure \cite{9298575} including fine-tuning all the layers for colorectal tissue classification. To improve the model training, different standard data augmentation techniques are employed, such as horizontal and vertical flips. However, we do not employ colour and geometry-based augmentations as they are observed to be sub-optimal in low data regimes. The randomness of the image sampling leads to large relative data variations compared to the number of images and, therefore, may harm the learning process. 

\noindent\textbf{Local Image Branch.}
This branch strives to capture local texture features complementary to the features learned in the standard global branch. The branch includes the same shared weights as those of the pre-trained ResNet model used in the standard global branch, except that now random samples of different sizes are fed to the model instead of the entire image. 
Specifically, in order to capture diverse texture patterns, we randomly sample patches ranging between 10-50 \% of the original image dimensions.
Next, we resize all the sampled local regions to a fixed size followed by the same set of data augmentations employed in the standard global branch.
Consequently, motivated by \cite{deit} in transformers, we employ a knowledge distillation mechanism by utilizing the corresponding output logits of both the standard global and local image branches. In this way, the network is forced to focus on local texture patterns that can likely aid colorectal tissue classification in low data scenarios. \\
\noindent\textbf{Loss Function.}
For the standard global branch, we employ the conventional cross-entropy loss \cite{NEURIPS2018_f2925f97} for classification, denoted as $\mathcal{L}_{main} = \mathcal{L}_{CE}(\hat{y}, y)$ in our architecture, where $\hat{y}$ is a predicted class and $y$ is a ground truth one-hot encoded label.
Furthermore, motivated by recent works in transformers and knowledge distillation \cite{deit},~\cite{caron2021emerging} and \cite{https://doi.org/10.48550/arxiv.2206.08491}, we perform self-distillation with a distillation loss $\mathcal{L}_{dist}$.
We utilize a shared weights model to achieve comparable image classification performance for both the entire image (standard global branch) and its smaller sampled region containing local texture patterns.
To this end, for knowledge distillation, we use the standard global branch as the teacher and the local image branch as the student. The teacher's output, being based on more data, is expected to be more confident and can be treated as a label.
In our approach, a "hard" label coming from the teacher branch is used as a target. Similar to \cite{deit}, the label is obtained by feeding a full image into the standard global branch, then taking the teacher's prediction $y_t = argmax(Z_t)$, where 
$Z_t$ is the output logits of the teacher after the $softmax$ function $\psi$.
This parameter-free type of label has previously been shown to provide better distillation performance.
Similarly, for the local image branch (student), we obtain its output logits $Z_s$ for the randomly sampled local image regions.

Next, depending on the amount of the available data per class $n_{im/c}$, we compute the distillation loss.
We observe that under extremely low data regimes (fewer than 20 images per class), the randomly sampled images may be of inconsistent complexity since adequate image diversity is not preserved. This situation may cause class complexity imbalance, where some classes have varying enough images to cover most of the important patterns, while others do not have this property.
Therefore, in order to mitigate this problem, for the sampled dataset portions with fewer than $n_{min} = 20$ images (usually $\leq 5 \%$ of the original dataset), instead of the standard cross-entropy loss we utilize focal loss \cite{focal_loss}, which is specifically designed to deal with the data imbalance.
In this way, the distillation loss can be described as follows:
\begin{equation}
    \label{eq:dist_loss}
    \mathcal{L}_{dist} =  
    \begin{cases}
        \mathcal{L}_{focal} (\psi (Z_s), y_t),  & \text{if } n_{im/c} \leq n_{min} \\
        \mathcal{L}_{CE} (\psi (Z_s), y_t),  & \text{otherwise,}
    \end{cases}
\end{equation}

Consequently, the final objective function $\mathcal{L}$ is defined as a combination of both classification and distillation losses:
\begin{equation}
    \label{eq:final_loss}
   \mathcal{L} = \frac{1}{2} \mathcal{L}_{main} + \alpha \frac{1}{2} \mathcal{L}_{dist} ,
\end{equation}
where $\alpha = 0.1$ is a controlling scaler for the self-distillation loss needed to manage the amount of complementary refinement information.

\section{Experiments and Results}

\begin{table}[t]\centering
\scriptsize

\caption{Comparison of different approaches using different percentages of the data on the Kather-2016 dataset. We conduct the experiments with three different train sets (for 1 \% to 50 \% splits) and report the mean classification accuracy. Our proposed KD-CTCNet achieves consistent improvement in performance over other methods across different data settings. Best results are highlighted in bold.}

\begin{tabular}[\textwidth]{l|c|c|c|c|c|c|c}
    \toprule
    Model & 1$\%$ & 3$\%$ & 5$\%$ & 10$\%$ & 20$\%$ & 
    50$\%$ & 
    100$\%$\\
    
    \midrule
    \tiny{ResNet-50}        &      72.64      &        82.88   &        86.29   & 88.91          &     91.82     &          
    94.57 &                    
    96.04 \\

    \tiny{ResNet-50 +  
    Sampling}      &      72.75      &        82.74   &        84.98   & 90.35          &     92.87     &          
    95.12 &                     
    96.65 \\

    \tiny{SAM ResNet-50} \cite{shu2022improving}        &      71.85      &        82.91   &        86.18   & 90.06          &     91.85     &                 
    94.65 &                     
    95.41 \\
    
    \midrule
    \tiny{\textbf{KD-CTCNet (Ours)}} &  \textbf{72.91\tiny{$\pm0.69$}} & \textbf{83.76\tiny{$\pm2.35$}} & \textbf{86.82\tiny{$\pm0.47$}} & \textbf{91.58\tiny{$\pm0.65$}} & \textbf{94.13\tiny{$\pm0.54$}} & 
    \textbf{95.76\tiny{$\pm0.15$}}  & 
    \textbf{96.88 
    } \\

    \bottomrule
\end{tabular}
\label{tab:ablation}
\end{table}

\begin{table}[t]\centering
\scriptsize
\caption{Performance of KD-CTCNet on NCT-CRC-HE-100K dataset.}
\begin{tabular}[\textwidth]{l|c|c|c|c|c|c|c}
    \toprule
    Model & 1$\%$ & 3$\%$ & 5$\%$ & 10$\%$ & 20$\%$ & 
    50$\%$ & 
    100$\%$\\
    
    \midrule
    \tiny{ResNet-50} & \textbf{73.1
    } & 86.66
    & 91.47
    & 94.10
    & 95.65 
    & 97.4
    & 97.8 \\
    \midrule
    \tiny{\textbf{KD-CTCNet (Ours)}} & 72.25\tiny{$\pm3.53$} & \textbf{87.21\tiny{$\pm1.98$}} & \textbf{92.67\tiny{$\pm1.25$}} & \textbf{94.94\tiny{$\pm0.74$}} & \textbf{96.31\tiny{$\pm0.37$}} & \textbf{97.91\tiny{$\pm0.22$}} & \textbf{98.23} \\

    \bottomrule
\end{tabular}
\label{tab:kather100}
\end{table}

\subsection{Implementation Details}
To adapt to low data regime settings, we apply the following changes.
First, unlike the default ImageNet resizing dimensions, images are resized to $192\times192$. This choice of dimensions is a trade-off between the optimal $224\times224$ size the model was pre-trained on and the original $150\times150$ size of the input images. Considering the receptive field of the ResNet-50 last convolutional layer, we select the closest number divisible by $32$ pixels size. We observe a slight deterioration in classification accuracy when using the $224\times224$ size.
To preserve the learnable texture, we do not use any colour or geometry-disturbing augmentations and only utilize horizontal and vertical flips.
For our local image branch, after performing random local image sampling from the input image, we resize the sampled regions to $96\times96$ pixels; a trade-off between the sizes of the standard ResNet-50 input and the average of our final truncated images.
The final choice is also selected as the closest value divisible by $32$.
Lastly, we perform the same horizontal and vertical flip augmentations here.
Our training setup includes the standard SGD optimizer with a momentum equal to 0.9, a learning rate of 0.01, and a training batch size of 32.
We emphasise that the choice of the above-mentioned hyper-parameters is geometrically-inferred and not optimised based on the test set, while the value for hyper-parameter $\alpha$ in \ref{eq:final_loss} is based on experiments performed on independent random train and test sets from the Kather-2016 dataset.
All experiments were conducted on a single NVIDIA RTX 6000 GPU using the PyTorch framework and the APEX utility for mixed precision training.

\subsection{Results and Analysis}

\noindent\textbf{Quantitative Results.}
Results in Table \ref{tab:ablation} demonstrate that our local image sampling strategy boosts the performance of the vanilla network across all sampled percentages, showing that the local sampling of the image introduces further local information to the network. 
Meanwhile, the SAM approach proposed for natural images does not perform well on CRC patches and its performance appears sub-optimal to the medical data domain.
Training KD-CTCNet from scratch yields much lower accuracies than those obtained by the other approaches, since the initialization of the other networks is based on their pre-trained weights, allowing the model to reach better results even in cases where data is scarce. However, using a pre-trained model in KD-CTCNet along with our local image sampling strategy proves to be superior to all other approaches across all data regimes reaching up to \textbf{2.67\%} improvement using only 10\% of the data. This proves the ability of our approach to capture better texture information. 
To showcase the scalability of KD-CTCNet, we also conduct experiments on another CRC dataset \cite{kather2019predicting} reported in Table~\ref{tab:kather100}. We observe that our KD-CTCNet improves over vanilla ResNet in almost all data percentages, proving its ability to enhance classification accuracy even in very low data regimes.

\begin{figure}[!h]
    \centering
    \includegraphics[width=1.0\textwidth]{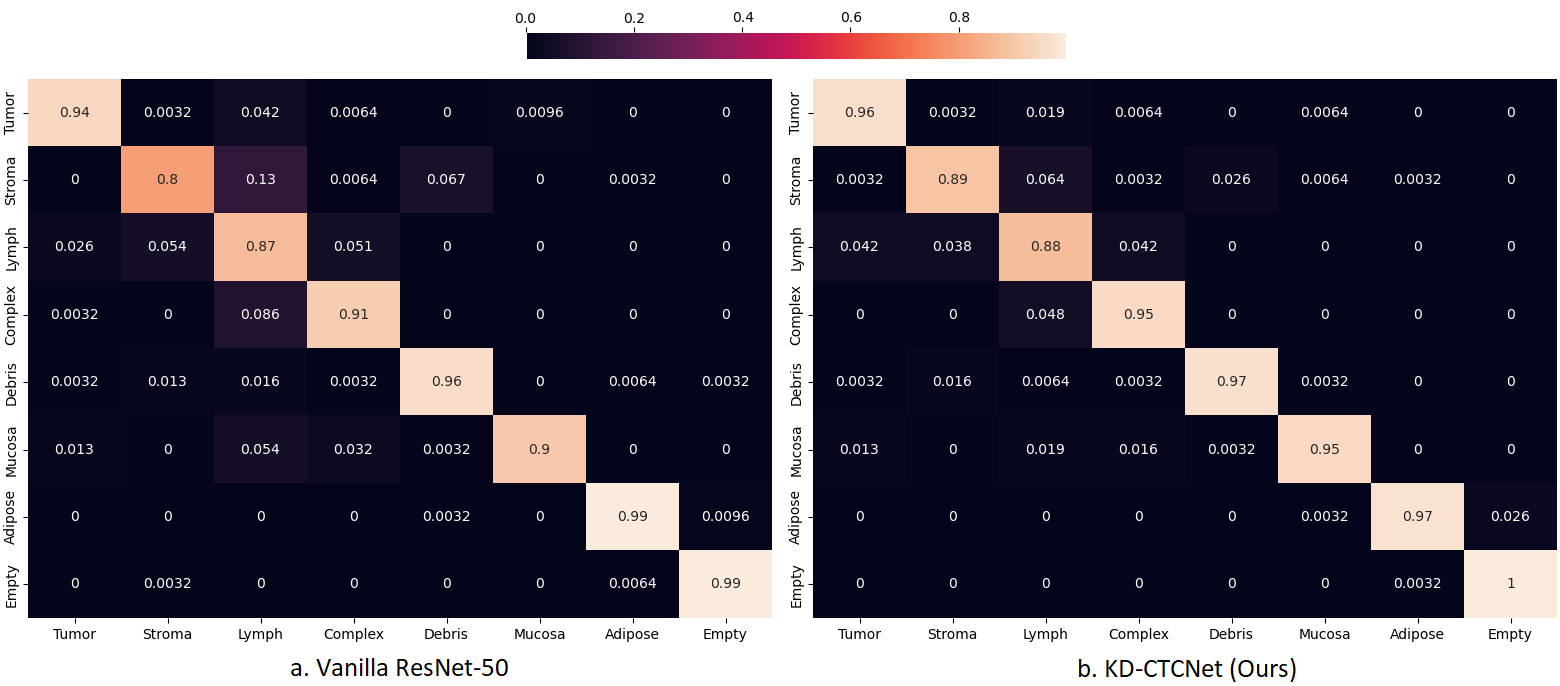}
    \caption{Comparison of confusion matrices calculated on the test set with 20 \% of the available data for (a) vanilla ResNet-50 and (b) our KD-CTCNet approach.}\label{fig:conf_m}
\end{figure}

\noindent\textbf{Qualitative Analysis.}
For qualitative analysis, two confusion matrices for the standard ResNet-50 model and our proposed KD-CTCNet are generated for one of the low data regimes (20 \%).
As can be seen in Figure~\ref{fig:conf_m}, our KD-CTCNet outperforms the vanilla ResNet in correctly classifying positive samples into their correct labels for most classes. And while ResNet struggles more in the distinction between stroma, lymph, and tumor, our model shows a higher discriminative ability in such cases, demonstrating the effectiveness of our approach in the CRC classification setting. On the other hand, our model seems to be less discriminative for adipose classification compared to the vanilla ResNet.

\section{Conclusion}

CRC histopathological image classification poses a great challenge for deep learning models, specifically in scenarios where data is scarce. In this paper, we propose a dedicated architecture for the classification of CRC scans and demonstrate its superiority to similar approaches.
More specifically, through extensive quantitative experiments, we showcase the ability of our model to reach advanced performance in the low data regime. This is achieved by enriching the standard CNN features with effectively captured local texture information from tissue samples through our specifically designed distillation loss.
Moreover, from the qualitative experiments, we show that learning low-level texture information allows the model to achieve better per-class accuracy and decrease confusion of visually similar classes by identifying subtle but important details.

\bibliographystyle{splncs04}
\bibliography{ref}

\end{document}